# Computing Reference Classes

Ronald P. Loui
Depts. of Computer Science and Philosophy
University of Rochester
Rochester, NY 14627

For any system with limited statistical knowledge, the combination of evidence and the interpretation of sampling information requires the determination of the right reference class (or of an adequate one).

The present note (1) discusses the use of reference classes in evidential reasoning, and (2) discusses a first implementation of Kyburg's rules for reference classes, and what we now know about implementing such a system in the future. This paper contributes the first frank discussion of how much of Kyburg's system is needed to be powerful, how much can be computed effectively, and how much is philosophical fat.

## 1. Reference Classes.

AI discussions on probability have perenially revolved around two problems: what to do with conflicting evidence, and how to get by without a lot of objective statistical knowledge. Each of these problems can be addressed by an adequate theory of how reference classes are selected.

Hans Reichenbach left modern philosophers of probability with a single task: in order to determine an event's probability, determine the narrowest reference class to which the event belongs, and about which adequate statistics are known [Rei49]. Suppose I know about the next Mets game, "m", that it is one in which Dwight Gooden will pitch, "Dm", and one to be played at home, "Hm", and one in which Keith Hernandez will bat, "Km"; I want to know the probability that the game will be a Mets' victory, P("Vm"). I have statistics about (or have an expert's degree of belief in) the per cent of Mets home games that are Mets' victories,

"%({x : Hx}, {x : Vx})", or just "%(H, V)",

and the per cent of Dwight Gooden home games that are Mets' victories,

"%(H & D, V)",

but no statistics (or beliefs) relating Keith Hernandez games to victories, whether at home or not, whether in conjunction with Dwight Gooden or not.

Some are willing to supply the missing numbers,

e.g., "%(H & D & K, V)",

by procedure or by fiat. But A.I. has left the age when inventing such numbers was condoned. So something must be done with the statistics that are legitimately known.

According to Reichenbach and his followers, if "%(H & D, V)" is adequately precise, say a narrow interval, then it gives the probability of the win. This is because "H & D" defines a narrower class than "H". Statistics for the narrowest class may not be available. Suppose "m" is an "M" game, "Mm" is true, where "M" is the predicate that individuates "m": i.e., "games identical to m". "M" could be analyzed into, say, the open sentence,

"H & D & M1 & M2 & ..."

Then "M" defines the narrowest class to which m belongs. But "%(M, V)" is surely not known, or else P("Vm") is a trivial query. We don't usually want the probability to come from statistics on V's among M's. Rules are needed to point out that only the "H & D" part of "M" is useful here. Systems of probability have been constructed on such rules.

Reichenbachian rules say what to do with evidence when it is conflicting or incomplete. When appropriate, the rules mandate combination of evidence via purely set-theoretic axioms. Set theory allows the construction of "%" statements from other "%" statements. At other times, the rules throw out information that is simply irrelevant. Discarding irrelevant information is natural in evidential reasoning though it is suppressed by the applied statistician in practice. Rules governing this practice need to be made explicit.

There is a close analogy to Bayesian methods. If

"Hm & Dm & Km"

is the total evidence, then P("Vm") is given by

P("Vm" | "Hm & Dm & Km").

Suppose, however, that this conditional is not known, and that what is known is just

P("Vm" | "Hm & Dm") and P("Vm" | "Hm").

Then some logical principles should determine that the total relevant, or total useful knowledge is "Hm & Dm". These logical principles should say what to do when there is some knowledge about P("Vm" | "Hm & Dm & Km") when it is only poor knowledge, such as belonging to the interval [.1, .9].

Even if the Bayesian uses point-probabilities, there is the matter of interpreting sampling information. Some Bayesians feel they should base their probability judgements not only on opinion, but on prior opinion modified by the experience of sampling. Sampling leads to k victories out of n Mets games, which leads to a posterior distribution on the probability of a victory. But which sampling? There are the samples of (a) Mets games played at home, (b) Mets games pitched by Dwight Gooden at home, and (c) Mets games in which Keith Hernandez went to bat. There may be no games in which Gooden pitched, Hernandez went to bat, and the Mets were at home. What if there was only one such game? There are no doubt differences in the amount of data in each sample; hence, different samples lead to posteriors based on different amounts of experience. If the Bayesian values experience, the choice matters.

Before asking in knee-jerk fashion what are to be the "weights for combination," some have asked what are the logical grounds for choosing among them. If the Mets are different from last year, last year's record doesn't matter. If they're on a winning streak, the relevant sample information should again be restricted. What would reflect this in the data? There may be Bayesian answers to such questions, but they are answers to the question of which reference class to use.

2. Kyburg's Strategy and Its Capabilities.

Henry Kyburg based his theory of probability on the determination of reference classes (so did John Pollock [Pol83, 84]). Kyburg's definitions first appeared in 1961 as a solution to the problems that led Carnap to Bayesianism. So it cannot be classified under Peter Cheeseman's taxonomy [Che85] as "invented new formalism." [Kyb74] is the complete presentation. [Kyb83] is a more cogent formulation.

2.1. The Idea.

The system follows Reichenbach, except that it uses interval-valued statistics for classes. A class is better than another if it is narrower, unless its interval contains the interval associated with the other class. So if

%(H & D & K, V) is [0, 1] and %(H, V) is [.3, .5],

the latter is better. Nested intervals are supposed to signal agreement; overlapping or disjoint intervals are supposed to signal disagreement.

Note that Kyburg just doesn't intersect or take unions of overlapping intervals in order to resolve their disagreement. If

%(H & D & K, V) is [.1, .6] and %(H, V) is [.4, .8],

the former is better, because its class is narrower. Between

%(H, V) = [.4, .8] and %(K, V) = [.3, .7],

neither is better. Since there is disagreement, neither is useable. Another class must be sought.

In order to determine the probability of t, a sentence in a formal language, relative to a base of knowledge,

1. Find statements of the form: "t ≡ (x ∈ Z)";
   i.e., isolate the events relevant to determining the probability in question;
   e.g., "I win bet b ≡ m ∈ the set of Mets victories".

2. Now for each sentence "x ∈ Y",
   i.e., for some property Y of x,
   e.g., "m ∈ the set of home games: {x : Hx}"
   find the strongest statistical statement for property Z among class Y: "%(Y, Z) = [p, q]";

   A statistical statement is stronger than another if the first's [p, q] interval is nested in the other's.
   These are the potentially useful statistics.
   Y's are "candidate" reference classes.

3. The 4-tuple <x, Y, Z, [p, q]> is an "inference structure for t."

4. Collect all such inference structures and call the set S.

5. Find IS*, the strongest member of S (the one with the strongest statistics) that "dominates" every member of S that "disagrees" with it (these are defined below).

6. If IS* is <x*, Y*, Z*, [p*, q*]>, then Prob(t) = [p*, q*].

Inference structures disagree when their [p, q]-intervals don't nest. IS1 dominates IS2 when IS1 "reflects" IS2 but IS2 doesn't reflect IS1. IS1 can reflect IS2 in any of a number of ways. The most interesting way is for the class of one to be a subset of the class of the other. If the statistics for Gooden games at home disagree with the statistics for games at home, then subset reflection will allow the former statistics. Subset reflection takes the most specific class when there is disagreement.

2.2. The Achievement.

Here's the achievement. Distinguish probability assertion,

Prob("x ∈ Z") = [p, q],

from statements about specific frequencies in classes,

"%(Y, Z) = [p, q]".

Then provide rules for selecting among such frequency statements in order to determine probability. Whatever information used to be used to determine or

184

combine probabilities, subjective or objective, can now be used to determine or combine frequencies. But where there is conflict there is no special problem. If one source of conflict reflects the other, there is deference. If there is no reflection, appeal is made to a third source, usually one that permits only a weaker conclusion. Where there is lack of information, again there is no special problem. The strongest permissible conclusion is drawn relative to the accepted knowledge.

2.3. Specific Constructions.

If "%(H, V)" and "%(K, V)" are known, but not "%(H & K, V)", some bounds on the latter can be constructed with computational effort, if there is knowledge of "%(H, K)", "%(K, H)", and so forth. The problem in general is an extremization of a non-linear objective with linear constraints.

But even if %(H & K, V) can't be determined to a narrow interval, there are other reflecting classes.

When the knowledge base contains:

"%(H, V) = $[p_1, q_1]$" and "%(K, V) = $[p_2, q_2]$".

it should also contain:

"$x \in V \equiv <x, x> \in V \times V$"; "$<x, x> \in H \times K$";

and "%(H × K, V × V) = $[p_1 p_2, q_1 q_2]$",

i.e., an inference structure based on

H × K. (ISX)

The product class is always a candidate inference structure, and it embeds a kind of independence assumption. It reflects the inference structures based respectively on H and on K.

It will almost always be dominated, in turn, by the inference structure based on the class:

"$\{<x, y> : x \in H$ and $y \in K$ and $x \in V \equiv y \in V\}$". (ISXB)

And the per cent of these that are in V × V is

  $[g(p_1, p_2), g(q_1, q_2)]$;

  $g(x, y) = xy/(1 - x - y + 2xy)$,

which is easily calculated. Using this inference structure would also be like making a kind of independence assumption (in this system, a provably better one, when evidence about the same event is combined).

But it too can't be used unless it dominates all others with which it disagrees, including whatever is known about joint information:

"$\{<x, y> : x \in (H \& K)$ and $y \in (H \& K)$ and $x = y\}$".

In general, interference could come from any structure of the form

"$\{<x, y> : x \in H$ and $y \in K$ and $x \in W \equiv y \in W...\}$"

where V is not a subset of W. For instance,

"$\{<x, y> : x \in H$ and $y \in K$

  and $x \in (V \& K) \equiv y \in (V \& H)\}$"

could interfere if its statistics were known to disagree.

We've identified only the inference structures based on (ISX) and on (ISXB) which reflect structures based on H and on K, which have statistics that are easily computable from "%(H, V)" and "%(K, V)". The one based on (ISXB) always dominates the one based on (ISX), so it is most interesting when there is conflict, and weak joint information.

3. Lessons from Implementation.
3.1. So Much Set Theoretic Computation.

> Lesson:
> The better you are with set theory, the better you'll
> be at computing reference classes.

Finding candidate reference classes and determining reflection are by far the major computations. Both are intensively set-theoretic.

3.1.a. Finding Classes.

Finding candidate classes starts with chaining on biconditionals. We implemented the biconditional chaining, but in practice, it's dispensable logical convenience. We want the probability of t. The equality reasoner leads to a set of pairs $<x_i, z_i>$, such that for each i,

  (IFF t (MEMBER $x_i$ $z_i$)) is true.

For each of these $x_i$, we have to find the sets to which it is known to belong. There is a choice (we implemented the latter, but now prefer the former):

a) chain forward on those sets to which $x_i$ is known to belong, and their supersets, $y_j$. Then either look for statistical statements

  (% ($y_j z_i$) (p q)), i.e., %($y_j z_i$) = [p, q],

or construct them.

b) chain backward to try to prove

  (MEMBER $x_i$ $y_j$)

for each $y_j$ whose $z_i$ statistics are known. This method requires that all inferred statistics be done by forward chaining. For those classes, e.g. $y_k$ and $y_l$, found to contain $x_i$, try to discover reflection among them by attempting proofs of the form

  (SUBSET $y_k$ $y_l$).

3.1.b. Computing Reflection.

Then there is the matter of reflection. Determining reflection is either determining one set is a subset of

185

another, or chaining on statistically equivalent inference structures, looking for subsets. Again, the computation is intensively set-theoretic.

With educated algorithmic design, it's possible to avoid having to compute some reflection relations. Deciding disagreement of inference structures is a much cheaper computation than trying to decide reflection. If IS1 and IS2 are inference structures, IS1 disagrees with IS2, and IS1 reflects IS2, then IS2 can't possibly be the reference class. So at the very least:

```
S is the set of inference structures, obtained under (b).
CHOICES ← S.
CHOICES is a priority queue ordered by strength.
Until CHOICES empties do {
    Select s* from CHOICES
    (s* is maximally strong among CHOICES).
    For every s ∈ S that disagrees with s* do {
        If s* reflects s
        then {
(i)         CHOICES ← CHOICES - s;
(ii)        if s reflects s* then exit with FAILURE.
        }
        else exit with FAILURE.
    }
    If not(FAILURE) then exit.
}
Take statistics from s*.
```

The deletions in (i) are the improvement over the naive algorithm. The line, (ii) can also be omitted for most purposes; reflection is anti-symmetric in real knowledge bases.

### 3.1.c. Reflection without Computation.

We can now explain why (a) seems more attractive than (b). Even with the shortcuts for computing reflection, there are too many proofs attempted. Instead, chain forward through known inclusion relations using one's favorite set-theoretic axioms. For every set found, one has simultaneously determined a candidate class and identified which other classes it reflects, i.e., which sets are its supersets.

The relations between inference structures direct the search, rather than lead to more proofs.

### 3.2. Reducing the Language.

*Lesson:*
*Set-theoretic language just gets in the way. The rules are too general: use a subset of the rules and restrict the languages over which probability is computed.*

The set-theory that is needed is significantly less elaborate than what Kyburg envisions, or what Brown [Bro78] or Pastre [Pas78] have provided for, in the past.

### 3.2.a. Restricting the Set-Forming Operations.

It is important to represent and manipulate sets like $\{x : x \in H \text{ and } y \in K \text{ and } x \in V \equiv y \in V\}$.

But for all the effort required to implement this set-theoretic structure faithfully, it is quite inexpensive to supply the relevant information manually. In fact, rather than name the set with a complex expression that shows its structure, in practice we found it easier just to give it a basic name and assert the relevant relations to other sets.

We found we could get along with a quite restricted language, where all sets are named by creative intersection. The details are not important. What is important is that the constructions from section 2.3, and the Bayesian constructions can be named with less set-theoretic machinery.

> EMORPH is the union of
> $\{<x, y> : x = y\}$, and
> $\{<x, y, z> : x = y = z\}$, etc.

(so it acts like it has polymorphic dimension on intersections; "E" for "equality").

> (BMORPH V) is also a union, of
> $\{<x, y> : x \in V \equiv y \in V\}$, and
> $\{<x, y, z> : x \in V \equiv y \in V \equiv z \in V\}$, etc.

then

> (I (X a b) EMORPH) is
> $\{<x, y> : x \in a \text{ and } y \in b \text{ and } x = y\}$,

And

> (I (X a b)(BMORPH V)) is
> $\{<x, y> : x \in a \text{ and } y \in b \text{ and } x \in V \equiv y \in V\}$.

With this kind of naming, all of the classes we've identified as important can be named and manipulated easily. A simple reasoner about inclusions, such as a type reasoner, will suffice if it can be made to handle multidimensional objects, intersections, and some sophisticated functions like those of functional programming languages.

### 3.2.b. A Minimal Language.

An even starker language we've considered for applications uses no asserted subset relations. No two sets with primitive names are known to be subsets. Two sets are subsets only if one is syntactically the intersection of the other with some additional sets. In this language, we represent only the classes, their intersections, and their "BMORPH" combinations: i.e., the undominated independence class with computable statistics, that look like (ISXB) discussed in section 2.3.
Define the function extern-prod. Let

(extern-prod( (a b c d) ) V ),

or just [abcd], be

{<x, y, z, t> : x ∈ a and y ∈ b and z ∈ c and t ∈ d and
x = y = z = t}.

Also let

(extern-prod( (a b c) (d) ) V ),

or just [abc,d], be

{<x, y, z, t> : x = y = z and z ∈ V ≡ t ∈ V and
z ∈ (a & b & c) and t ∈ d}.

In the notation that uses only bracket and comma, V is implicit. [abc,d] is the class that would be formed if one took the joint information about the class a & b & c, and did the (ISXB)-style product with the class d.

In this language, it's very easy to see which classes to appeal to when there is conflict, and what are the reflection relations among inference structures built on these classes. Suppose x ∈ [abcd]. For the probability of x in V, appeal first to %([abcd], V). If there is knowledge about %([abc], V) and %(d, V), then compute %([abc,d], V).

Consider using it if it is stronger than %([abcd], V). If it is, then [abc,d] must also agree with all that are computable among: [ab,cd], [abd,c], [ac,bd], [acd,b], [ad,bc], [bcd,a], [bd,a,c], and so forth. The only classes that [abc,d] is allowed to disagree with are those reflected classes such as [ab,c,d] and [ab,c], which can be determined by the following easy rule. Reflected classes are composed of constituents, such as "ab", and "c", and "d". And for each constituent, e.g., "ab", there is some constituent of [abc,d], "abc", which contains all of the first constituent's characters (set names).

4. Concluding Discussion.

Kyburg's system does solve some perplexing questions inherent in evidential reasoning, including: 1) when does an independence assumption conflict with joint information; 2) which conclusion can be drawn, if any, when there is disagreeing evidence; and 3) how to choose between evidence that is strong but ill-founded, and evidence that is well-founded but weak.

The key elements are: 1) the use of intervals, so that probability assertions can vary in strength, and 2) the distinction between frequency and probability assertion, so that conflict among statistical information can be arbitrated outside of the probability calculus.

We implemented the system in some generality, and found that the difficulties were in deciding how much set-theoretic language and how much set-theoretic inference to use.

In order to simplify, we studied exactly which inference structures would be useful, of the many that satisfy the definitions. In particular, we know that when there is conflict between two classes, e.g., %(H, V) and %(K, V), one should appeal to %(H & K, V) if it is known, or else to %( [H, K] , V), if it is agreeable with %(H & K, V).

A satisfactory general inference engine could be based on the language of a multidimensional type reasoner, and applications could be successfully based on the even starker language, which makes all reflection relations obvious by inspection.

In Kyburg's favor, the system copes with the traditional problems as advertised. In his disfavor, the system does not have much locality; bits of knowledge in all sorts of corners can be relevant, and are not easily pre-processed. So it is promising for diagnosis and identification problems where time is not as important as error, where carefully integrating evidence is more important than processing lots of information quickly.

The next step is to see how well these ideas fare in applications.